%% file: main.tex
\documentclass{article}

 \usepackage{xspace}
\newcommand{\eg}{\emph{e.g.,}\xspace}
\newcommand{\ie}{\emph{i.e.,}\xspace}

 \usepackage[main, final]{neurips_2026}

\usepackage[utf8]{inputenc} 
\usepackage[T1]{fontenc}    

\usepackage{graphicx}
\usepackage{booktabs}
\usepackage{multirow}

\usepackage{hyperref}       
\usepackage{url}            
\usepackage{booktabs}       
\usepackage{amsfonts}       
\usepackage{nicefrac}       
\usepackage{microtype}      
\usepackage{xcolor}         
\usepackage{amsmath}
\usepackage[nameinlink]{cleveref}
\usepackage{mathtools}
\usepackage{enumitem}

\crefname{section}{Section}{Sections}
\crefname{figure}{Figure}{Figures}
\crefname{table}{Table}{Tables}
\crefname{appendix}{Appendix}{Appendices}

\title{CD-RCM: Generalizable Continuous-Depth Novel View Synthesis for Reflectance Confocal Microscopy}

\author{
Tooba Imtiaz$^{1}$, Milind Rajadhyaksha$^{2}$, Kivanc Kose$^{2,*}$, Jennifer Dy$^{1,*}$\\
$^1$Northeastern University, 
$^2$Memorial Sloan Kettering Cancer Center\\
\texttt{\{imtiaz.t,j.dy\}@northeastern.edu},
\texttt{\{rajadhym,kosek\}@mskcc.org}\\
\textnormal{\footnotesize $^*$~denotes equal contribution}
}

\begin{document}

\maketitle

\begin{abstract}
Reflectance confocal microscopy (RCM) provides noninvasive, cellular-resolution “optical biopsies” of human skin {\em in vivo} by acquiring en-face images at successive depths, forming a sparse $z$-stack. Due to optical limitations, these stacks are anisotropic 3D volumes with lateral resolution ($0.5\mu m$) $\sim$6 times higher compared to axial resolution, which is defined by the optical sectioning ($3\mu m$), limiting the interpretation of tissue. Our goal is to provide continuous-depth visualization by interpolating intermediate sections and making the 3D volume isotropic.
Such a representation permits arbitrary-direction sectioning, including histopathology-like cross-sectional examination, without requiring per-patient optimization.
To that end, we introduce the first RCM-specific novel-view synthesis (NVS) approach, CD-RCM, a feedforward model that predicts realistic, unseen depths from sparsely sampled RCM stacks.
Classical neural rendering methods focus on reconstruction from surface-level multi-view observations.
In contrast to surface-level camera views, RCM can acquire optically sectioned en-face images of tissue beyond the surface up to $200 \mu m$. However, during visualization of the RCM stacks, observations of the shallower sections (towards the surface) obscure the deeper ones. This unique axial imaging geometry and layer-dependent anatomical organization motivated our development of a tailored architectural and training framework that explicitly accounts for RCM’s depth-resolved, occlusive imaging physics.
Experiments demonstrate that CD-RCM achieves high-fidelity novel-view synthesis with sub-second inference time.
\end{abstract}

\input{sections/0-introduction_v2}
\input{sections/1-related-works_v2}
\input{sections/2-method_v2}
\input{sections/3-experiments_v2}
\input{sections/4-results_v2}
\input{sections/5-ablations_v2}

\input{sections/6-expert-eval}
\input{sections/7-limitations-and-discussion}
\input{sections/8-conclusion}

\section*{Acknowledgments}
The reading study was conducted by Dr. Miguel Cordova, an RCM imaging expert with approximately 20 years of experience acquiring and interpreting reflectance confocal microscopy images at the Dermatology Service of Memorial Sloan Kettering Cancer Center (MSKCC).

This work is supported by grants DoD/USAMRAA HT94252410553 and DoD/USAMRAA HT94252410554, NIH/NCI R01CA199673, NIH/NCI U24-CA264369, R01CA293974, and NIH/NCI Cancer Center Support Grant P30 CA008748.

\bibliographystyle{plainnat}
\bibliography{main}


\newpage
\appendix
\section*{Appendix}
\input{supp-sections/9-additional-implementation-details}
\input{supp-sections/10-dataset-audit}
\input{supp-sections/11-additional-ablation}
\input{supp-sections/12-eval-baselines}
\input{supp-sections/13-eval-metrics}
\input{supp-sections/14-additional-qual-results}

\end{document}

%% file: sections/0-introduction_v2.tex
\section{Introduction}\label{sec:intro}

Early detection and precise staging of skin disease rely on observing microstructural changes across depth, often beginning at anatomically meaningful boundaries such as the dermal–epidermal junction (DEJ). Optical sectioning microscopy techniques like Reflectance Confocal Microscopy (RCM)~\citep{gonzalez2008reflectance} enable noninvasive \emph{in vivo} imaging of skin with near-histologic resolution and are widely used for diagnosing tumors, inflammatory conditions, and guiding surgical margin assessment~\citep{edwards2017diagnostic, HUZAIRA2001846, jcm14165779, richarz2022challenges}.

Clinicians acquire RCM data as \emph{z-stacks} — sequences of en-face optical sections sampled at discrete depths. Although lateral resolution is high ($\sim$0.5 $\mu$m), axial resolution is governed by optical sectioning ($\sim$3 $\mu$m), resulting in strongly anisotropic volumes that limit depth-wise interpretation, particularly near diagnostically critical transitions such as the DEJ. Prior studies note the tradeoff between slice spacing and acquisition time, and the sensitivity of downstream algorithms to sampling density~\citep{bozkurt2017unsupervised, ghanta2016marked}. Conventional interpolation methods (\eg spline-based) lack anatomical priors and fail to preserve depth-dependent skin morphology.

In this work, we introduce \emph{Continuous-Depth Novel-View Synthesis for RCM} (CD-RCM), a feedforward model for continuous-depth novel-view synthesis from sparse RCM stacks. Our approach leverages the structured axial acquisition process of optical sectioning microscopy and models depth variation as a queryable dimension. Unlike prior methods, CD-RCM predicts anatomically consistent intermediate slices in a single forward pass, without per-stack optimization, and generalizes across patients and imaging sites. Rather than performing optical super-resolution, CD-RCM exploits the substantial overlap between adjacent sections, acquired every $1.5\,\mu$m relative to an axial section thickness of approximately $3$--$5\,\mu$m, to densify sampling along depth. 
It can estimate intermediate optical sections consistent with structures already resolved by RCM, but cannot recover sub-resolution structures or spatial frequencies absent from the acquired measurements. 
The resulting volume is more densely and isotropically sampled, while its intrinsic optical resolution remains unchanged.

To preserve fine cellular structure, we further propose a \emph{skin-specific perceptual loss}, based on a domain-adapted feature extractor trained on RCM data. This enables the model to capture micro-textural patterns that are not well represented by perceptual losses trained on natural images.

In summary, we make the following contributions:
\begin{enumerate}
  \item We introduce the first formulation of arbitrary-depth novel-view synthesis for RCM, explicitly addressing the anisotropic and see-through nature of RCM stacks.
  \item We develop CD-RCM, a feedforward, depth-queryable model that incorporates the axial imaging physics of RCM and generalizes across skin sites without per-stack optimization.
  \item Our experiments demonstrate that CD-RCM achieves high-fidelity novel-depth synthesis in a single inference pass, outperforming conventional interpolation baselines. In a blinded expert evaluation, CD-RCM received ratings comparable to acquired slices and was selected as the reconstruction most similar to ground truth in majority cases.
\end{enumerate}

%% file: sections/1-related-works_v2.tex
\section{Related Work}\label{sec:related-work}
Our work draws on two areas of prior research: algorithmic analysis of RCM stacks, and neural methods for 3D reconstruction and novel-view synthesis. We review each below, highlighting how existing methods address related problems but leave a gap at their intersection — namely, feedforward synthesis of arbitrary intermediate depths from sparsely sampled confocal stacks.

\subsection{RCM stacks and downstream analysis}

RCM enables non-invasive, {\it in vivo} optical sectioning of skin with near-cellular lateral resolution. Unlike conventional histopathology, which provides vertically sectioned, stained tissue spanning multiple layers, RCM produces grayscale {\it en face} images based on intrinsic reflectance contrast. The en face geometry and anisotropic resolution make interpretation challenging, as clinicians must mentally reconstruct cross-sectional structure from sparsely sampled depth stacks.

Prior computational work has focused on algorithmic analysis of discretely acquired stacks — including DEJ localization~\citep{kose2020detection, ghanta2016marked}, strata delineation, and layer thickness estimation~\citep{bozkurt2017unsupervised, bozkurt2021skin} — treating stacks as fixed collections of optical sections rather than modifying or densifying the underlying volume.

From a visualization standpoint, RCM's anisotropic resolution yields inadequate reconstructions from any angle other than en face. Perpendicular sections analogous to histopathology can be constructed from stacks but suffer from sparse axial sampling; na\"ive interpolation introduces blocking artifacts and unrealistic structures. Dense 3D visualization~\citep{3DPath} could enhance structural understanding of tissue organization, but alternative volumetric modalities such as light-sheet microscopy~\citep{LightSheetM} and LC-OCT~\citep{LCOCT} are either impractical for routine in vivo imaging or offer lower lateral resolution than RCM.

Crucially, each RCM optical section integrates signal from adjacent depths due to the finite axial point-spread function, yet this overlapping information has not been leveraged for dense axial reconstruction. We formulate continuous-depth reconstruction as a feedforward novel-view synthesis problem that exploits this property and the structured axial acquisition geometry to densify sparse stacks without needing per-stack optimization.

\subsection{Neural rendering, 3D reconstruction and view synthesis}

\begin{figure}
    \centering
    \includegraphics[width=\linewidth]{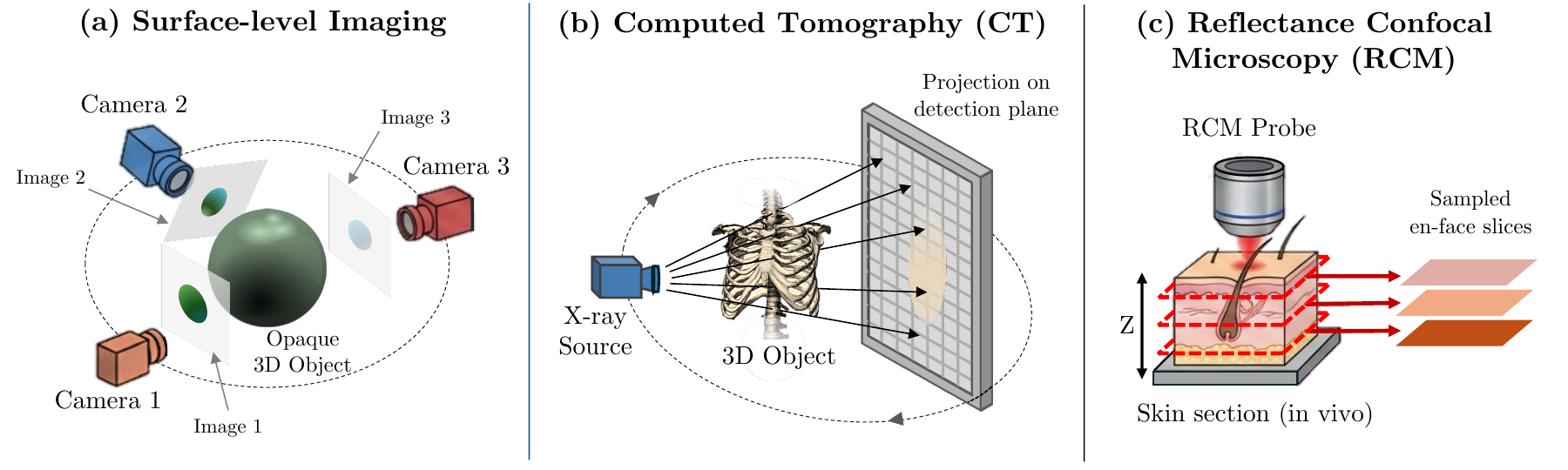}
    \caption{\textbf{Comparison of imaging paradigms.} 
    (a) Surface-level imaging reconstructs opaque objects from multiple posed viewpoints with geometric parallax. (b) CT acquires transmission projections from multiple angles and reconstructs 3D structure from X-ray attenuation. (c) RCM performs axial optical sectioning, capturing discrete en-face slices of internal tissue at successive depths.}
    \label{fig:imaging-comparisons}
    \vspace{-5mm}
\end{figure}

3D reconstruction and novel-view synthesis have progressed from implicit representations~\citep{mescheder2019occupancy, chibane2020neural} to NeRF~\citep{mildenhall2021nerf} and 3D Gaussian Splatting \citep{kerbl20233d}, enabling increasingly high-fidelity reconstruction with reduced optimization time.
However, two limitations persist. Most methods require per-scene optimization, and while recent generalizable frameworks~\citep{charatan2024pixelsplat, zhang2024gs, jin2025lvsm, imtiaz2025lvt} enable feedforward inference, they are primarily designed for surface-level reconstruction of opaque objects under known camera poses (\cref{fig:imaging-comparisons}a), offering limited support for internal geometry.

Tomographic reconstruction offers a parallel perspective, as recovering 3D structure from sparse measurements is an ill-posed inverse problem (\cref{fig:imaging-comparisons}b). Recent radiative Gaussian splatting approaches~\citep{cai2024radiative, r2_gaussian, li20253dgr} tailor rendering operators to X-ray physics for sparse-view CT, but like surface-based NVS methods, they require per-case optimization and do not generalize across subjects or hardware settings without retraining. Video frame interpolation~\citep{reda2022film} has also been adapted for intermediate-plane synthesis in 3D tomography~\citep{gambini2024video}.

As illustrated in \cref{fig:imaging-comparisons}, RCM occupies a distinct regime that differs fundamentally from both surface-level and CT imaging modalities. Unlike standard NVS, RCM captures internal tissue via axial optical sectioning with minimal viewpoint variation; unlike CT, its contrast is governed by depth-dependent tissue optics rather than geometric parallax. These properties motivate a specialized, generalizable approach for depth-conditioned synthesis of en face sections from sparse axial stacks.

Recent neural-field approaches have also addressed inverse problems in computational microscopy. \cite{cao2024neural} jointly recover scene dynamics and structure from multi-shot measurements, while \cite{kang2026adaptive} estimate optical aberrations and sample structure in two-photon microscopy. Unlike these per-sample, physics-based reconstruction methods, CD-RCM operates on already reconstructed optical sections and learns an amortized cross-stack model for continuous-depth synthesis.

%% file: sections/2-method_v2.tex
\section{Method}\label{sec:method}
CD-RCM builds upon transformer-based novel-view synthesis architectures~\citep{charatan2024pixelsplat, zhang2024gs, jin2025lvsm}, adapting them to model internal skin structures captured by reflectance confocal microscopy. We describe our components below and highlight key differences from conventional surface-level NVS frameworks.

\subsection{Overview}\label{sec:overview}
An RCM \emph{stack} consists of $N$ en-face grayscale images $\mathbf{I}_i \in \mathbb{R}^{H \times W}$, $i=1\ldots N$, each comprising a 3\,$\mu$m thick optical section of tissue. By moving the focusing depth through the tissue, the microscope translates the focal plane along the axial ($z$) direction at $>$1.5\,$\mu$m steps with negligible lateral ($x,y$) motion, imaging up to 150--200\,$\mu$m deep. Stacks typically contain 50--65 slices.

RCM does not operate as a conventional camera-based imaging system. Instead, it raster-scans a focused beam using galvanometric mirrors and collects back-scattered light through a confocal pinhole, which enforces optical sectioning and provides depth selectivity. Since each slice corresponds to a fixed lateral field of view at a known axial position, we approximate the imaging process as a virtual pinhole camera undergoing pure $z$-axis translation (\cref{sec:pose,sec:intrinsics}). This abstraction enables us to leverage Pl\"ucker ray embeddings and transformer architectures designed for multi-view geometry.

\begin{figure}
    \centering
    \includegraphics[width=\linewidth]{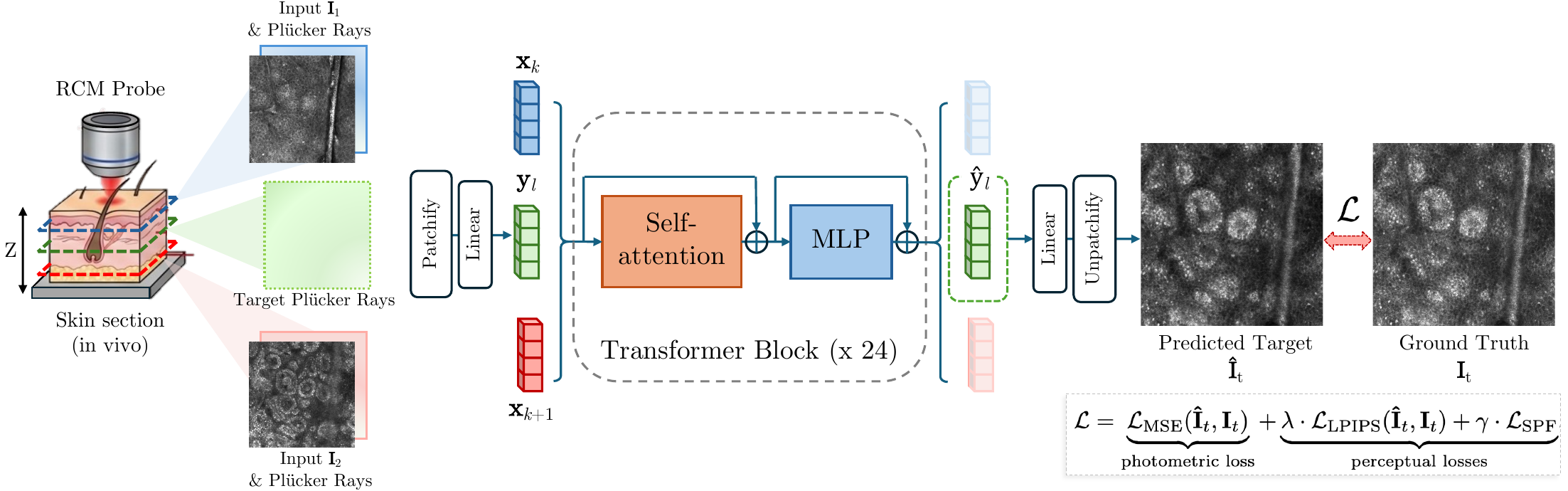}
    \caption{\textbf{Overview of CD-RCM.} Sparse input RCM slices and their Pl\"ucker ray embeddings are mapped to a unified token space and processed by a decoder-only transformer. Target ray tokens condition the synthesis; updated target tokens are decoded to produce $\hat{I}_t$. Training uses a weighted combination of photometric, LPIPS, and skin-specific perceptual feature loss $\mathcal{L}_\text{SPF}$.}
    \label{fig:overview}
\end{figure}

Given sparse inputs $\mathbf{I}_i$, CD-RCM synthesizes target slices $\mathbf{I}_t$ at arbitrary unseen depths (\cref{fig:overview}) within the range of the collected stack.
As illustrated in \cref{fig:overview}, we map input slices and their corresponding Pl\"ucker embeddings into a unified token space, process them using a decoder-only transformer, and decode only the updated target tokens to generate $\hat{\mathbf{I}}_t$. We train the model using a combination of photometric reconstruction loss and perceptual supervision, including our proposed skin-specific perceptual feature loss.

Below, we describe the virtual camera parameters (\cref{sec:pose,sec:intrinsics}), skin-specific perceptual feature extractor (\cref{sec:ss-perceptual-feats}), CD-RCM architecture (\cref{sec:arch}), and overall training objective (\cref{sec:loss}).

\subsection{Virtual Camera Parameters}\label{sec:pose}
RCM forms images via a scanning laser and confocal pinhole that optically section tissue at controlled focal depths. Although no physical camera parameters are available, the acquisition process provides strong geometric priors, \ie consecutive slices share the same lateral field of view and differ primarily in axial position.
Note that, unlike surface-level datasets, where pose estimation requires feature matching across viewpoints, RCM stacks exhibit near-perfect lateral alignment and systematic axial translation. Inter-slice correspondence arises from imaging different axial sections of the same underlying three-dimensional tissue structures (e.g., cells, vessels, and collagen), rather than from observing a static object under varying viewpoints. Consequently, the depth variation reflects progressive optical sectioning rather than geometric parallax. We exploit this by modeling acquisition as a virtual pinhole camera with fixed orientation translating along $z$.

Let slice index $i \in \{1, \dots, N\}$ denote the depth position within a stack. We approximate the camera motion as a pure translation, \ie $\mathbf{c}_i = [0,0,z_i]^T, z_i = i \cdot \Delta_z$.
Here $\Delta_z$ is a constant axial step size. In practice, $\Delta_z$ corresponds to the physical spacing between optical sections (approximately 3$\mu$m), mapped to normalized model coordinates.

Assuming fixed camera orientation, the world-to-camera transform is
\begin{equation}
\mathbf{T}_{\text{w2c}}^{(i)} =
\begin{bmatrix}
\mathbf{I}_{3} & \mathbf{c}_i \\
\mathbf{0}^\top & 1
\end{bmatrix}
\end{equation}
where $\mathbf{I}_{3}$ denotes identity rotation.
Neural rendering formulations typically adopt the camera-to-world convention. Since the transform is purely translational, its inverse admits a closed-form expression:
\begin{equation}
\mathbf{T}_{\text{c2w}}^{(i)} =
\begin{bmatrix}
\mathbf{I}_{3} & -\mathbf{c}_i \\
\mathbf{0}^\top & 1
\end{bmatrix}
\end{equation}

This matrix defines the camera pose associated with slice $I_i$.

\paragraph{Stack Canonicalization.}
Although poses are synthetically constructed, per-stack normalization improves numerical stability and learning consistency across stacks. Given a set of translations $\{\mathbf{c}_i\}$, we compute the mean camera center $\bar{\mathbf{c}} = \frac{1}{N} \sum_{i=1}^{N} \mathbf{c}_i$ and the maximum absolute translation $\mathbf{c}_\infty = \underset{i \in \{1,\dots,N\}}{\max}\|\mathbf{c}_i\|_{\infty}$ to standardize all poses via $\mathbf{c}_i \leftarrow (\mathbf{c}_i - \bar{\mathbf{c}}) / \mathbf{c}_\infty$.
This embeds each stack within a shared normalized coordinate system while preserving relative depth relations.

\paragraph{Camera Intrinsic Parameters}\label{sec:intrinsics}
Camera intrinsics $K$ are similarly approximated using a fixed pinhole model:
\begin{equation}
\mathbf{K} =
\begin{bmatrix}
f_x & 0 & u_0 \\
0 & f_y & v_0 \\
0 & 0 & 1
\end{bmatrix}
\end{equation}
where focal lengths are set proportional to image resolution, \ie
$f_x = f_y = \alpha H$,
with $\alpha$ a constant scaling factor, and $u_0 = v_0 = \frac{H}{2}$. Altogether, this approximation provides a stable geometric reference without requiring physical calibration.




\subsection{CD-RCM Architecture}\label{sec:arch}

Inspired by the decoder-only transformer design of LVSM~\citep{jin2025lvsm}, we train a feedforward depth-queryable model that renders intermediate slices between acquired slices. Unlike LVSM, which operates on surface views with geometric parallax, CD-RCM models purely axial variation and depth-conditioned reflectance.

\paragraph{Tokenizing Input Views.} For each input view $\mathbf{I}_i$, we compute pixel-wise Pl\"ucker ray embeddings~\citep{plucker1865xvii} $\mathbf{P}_i \in \mathbb{R}^{H\times W\times 6}$ from the virtual camera parameters, and tokenize both into non-overlapping $p \times p$ patches following ViT~\citep{dosovitskiy2020vit} to obtain $\mathbf{I}_{ij}$ and $\mathbf{P}_{ij}$, respectively. We channel-wise concatenate each pair, and map to $d$-dimensional input patch tokens $\mathbf{x}_{ij}$ via a linear layer:
\begin{equation}
    \{\mathbf{x}_{ij}\}_{j=1}^{HW/p^2} = \text{Linear}_{\text{input}}(\text{Concat}(\mathbf{I}_{ij}, \mathbf{P}_{ij})) \in \mathbb{R}^d.
\label{eq:token-ip}
\end{equation}
Lastly, we flatten these to obtain a 1D input token sequence $\mathbf{x}_K$ of length $K = NHW/p^2$.

\paragraph{Tokenizing Target Views.} For each target depth $t = 1,\ldots,T$, we similarly compute Pl\"ucker embeddings $\mathbf{P}_t$, patchify into $p\times p$ patches to obtain $\mathbf{P}_{tj}$, and map these to $d$-dimensional target tokens $\mathbf{y}_{tj}$ by a linear layer:
\begin{equation}
    \{\mathbf{y}_{tj}\}_{j=1}^{HW/p^2} = \text{Linear}_{\text{target}}(\mathbf{P}_{tj}) \in \mathbb{R}^d.
\label{eq:token-tar}
\end{equation}
We flatten these to form a 1D target token sequence $\mathbf{y}_L$ of length $L = THW/p^2$.

\paragraph{Transformer-based Decoder.} We concatenate $[\mathbf{x}_K, \mathbf{y}_L]$ and process the sequence with a decoder-only transformer comprising 24 self-attention blocks (with QK-normalization~\citep{henry2020query}) and MLPs. Here, the term \emph{decoder} refers to a GPT-style architecture~\citep{radford2019language} without causal masking, \ie attention is bidirectional across all tokens. After transformation, we retain only the updated target tokens and apply a linear layer followed by sigmoid activation to recover pixel intensities:
\begin{equation}
    \mathbf{\hat{I}}_{\ell} = \text{Sigmoid}(\text{Linear}_\text{out}(\mathbf{\hat{y}}_\ell)) \in \mathbb{R}^{p^2}
\end{equation}
Finally, we unpatchify (reshape) the intensity patches to recover the 2D synthesized target slices $\mathbf{\hat{I}}_t$.

\subsection{Skin-Specific Perceptual Feature Extractor}\label{sec:ss-perceptual-feats}

Standard perceptual losses~\citep{johnson2016perceptual, zhang2018unreasonable} emphasize natural-image semantics over the micro-textural patterns that dominate RCM imagery. To provide supervision sensitive to skin-specific structures (cellular morphology, speckle statistics, layer-dependent contrast), we train a domain-adapted perceptual backbone on a large auxiliary dataset of RCM mosaic crops.

We initialize a ViT~\citep{dosovitskiy2020vit} from self-supervised DINOv3~\citep{simeoni2025dinov3} and adapt it to the skin domain via LoRA~\citep{hu2022lora} applied to all attention projections:
\begin{equation}
\phi_s(\cdot) = \phi_{\text{DINOv3}}(\cdot; \theta_{\text{base}} + \Delta \theta_{\text{LoRA}}),
\end{equation}
where only $\Delta \theta_{\text{LoRA}}$ is optimized. A teacher network $\phi_t$ is maintained via EMA \ie $\theta_t \leftarrow m\theta_t + (1-m)\theta_s$, with momentum coefficient $m$. We follow the standard DINO multi-crop training paradigm~\citep{simeoni2025dinov3}, where we generate 2 global ($224\times224$) and 8 local ($96\times96$) crops per image. The teacher processes global crops, while the student processes all crops, promoting local–global consistency and robustness. For training, we use cross-entropy distillation between student and teacher distributions, computing losses from both CLS and mean-pooled patch tokens. Once trained, $\phi_s$ is frozen and used as a skin-specific feature extractor.

\subsection{Training Objective}\label{sec:loss}

We train CD-RCM using a weighted combination of complementary photometric and perceptual losses:
\begin{equation}
\mathcal{L} =
\underbrace{\mathcal{L}_\text{MSE}(\mathbf{\hat{I}}_t, \mathbf{I}_t)}_{\text{photometric loss}}
+ \underbrace{\lambda \cdot \mathcal{L}_\text{LPIPS}(\mathbf{\hat{I}}_t, \mathbf{I}_t)
+ \gamma \cdot \mathcal{L}_\text{SPF}}_{\text{perceptual loss}},
\label{eq:loss}
\end{equation}
where $\lambda$ and $\gamma$ are scalar weights. $\mathcal{L}_\text{MSE}$ computes pixel-wise mean squared error between $\mathbf{\hat{I}}_t$ and $\mathbf{I}_t$, enforcing structural accuracy. $\mathcal{L}_\text{LPIPS}$~\citep{zhang2018unreasonable} captures higher-order perceptual structure. Following~\citep{imtiaz2025lvt}, we use LPIPS over the VGG-based loss~\citep{johnson2016perceptual} of prior works~\citep{zhang2024gs, jin2025lvsm}. Our \emph{skin-specific perceptual feature} loss,
\begin{equation}
\mathcal{L}_\text{SPF} = \frac{1}{T}\|\phi_s(\mathbf{\hat{I}}_t) - \phi_s(\mathbf{I}_t)\|_1,
\label{eq:loss-spf}
\end{equation}
complements $\mathcal{L}_\text{LPIPS}$ by enforcing consistency in a feature space adapted to dermatologic imagery, encouraging preservation of diagnostically relevant morphological details.

\subsection{Training and Inference}\label{sec:training-pipeline}
During training, we sample sequences of $M$ consecutive slices from each stack, use the first, middle, and last as conditioning inputs $\mathbf{I}_i$, and randomly select four intermediate slices as targets $\mathbf{I}_t$. This randomized sampling across varying depth separations prevents overfitting to fixed spacing and enables generalization to arbitrary depths. At inference, CD-RCM takes any three evenly spaced slices and predicts target views at arbitrary intermediate depths in a single feedforward pass — no per-stack optimization, iterative refinement, or perceptual feature computation is required. Effectively, this allows synthesis at continuous axial positions, which, in practice, corresponds to generating non-integer-index slices between discretely acquired sections.

%% file: sections/3-experiments_v2.tex
\section{Experiments}\label{sec:experiments}
We describe our experimental setup (\cref{sec:exp-setup}) and discuss the results obtained (\cref{sec:results}). We also ablate key design choices in \cref{sec:ablation}.
\subsection{Experimental Setup}\label{sec:exp-setup}

\paragraph{Dataset.}
We conduct our experiments using a dataset comprising 216 RCM stacks acquired \emph{in vivo}. Each stack corresponds to a single skin site and contains between 50 and 65 en-face optical sections sampled along the axial ($z$) direction. Imaging was performed on healthy skin of volunteers without any special preparation, with images obtained from the arm and trunk of healthy volunteers aged 20–50 years. It is important to note that each stack originates from a distinct skin site, and even when acquired from the same subject, RCM images do not exhibit visual similarity across sites due to strong spatial and structural variability. As a result, stacks can be treated as independent samples. While subject-level metadata is unavailable, our train/test split is performed at the stack level and is effectively disjoint. 
We perform a post hoc audit for potential overlap between training and evaluation stacks reported in Appendix~\ref{sec:dataset_overlap}.

Data were collected using a commercially available reflectance confocal microscope (VivaScope 1500, Caliber Imaging and Diagnostics, Rochester, NY), a system widely used in clinical imaging and dermatologic studies~\citep{RCM_LSM, atak2023confocal}. The microscope employs a stabilized imaging window that adheres to the region of interest, ensuring a fixed lateral field of view during stack acquisition.

Each slice is a single-channel intensity image of resolution $1000 \times 1000$ pixels, corresponding to a lateral resolution of approximately $0.5\,\mu m$. The optical sectioning thickness (axial resolution) of the imaging system is approximately $3\,\mu m$, while consecutive slices are acquired at $1.5\,\mu m$ intervals. This results in axial oversampling relative to the optical sectioning limit, yet the stacks remain strongly anisotropic due to the disparity between lateral and axial resolution.

\paragraph{Preprocessing.}
We register each stack using SIFT-based~\citep{lowe2004distinctive} affine alignment implemented in Fiji~\citep{schindelin2012fiji} to reduce motion-induced inter-slice misalignment. After registration, we extract $960 \times 960$ center crops to remove boundary artifacts from the imaging window and resize to $512 \times 512$ via bilinear interpolation. The final five slices of each stack are excluded due to low signal content. We use a train-test split of approximately 70/30, yielding 156 training and 60 evaluation stacks.

\paragraph{Implementation Details.}
We train CD-RCM on four NVIDIA A6000 GPUs using a coarse-to-fine strategy~\citep{zhang2024gs,jin2025lvsm}. We first train CD-RCM from scratch for 20k steps at $256 \times 256$ resolution (batch size 8 per GPU), followed by 10k finetuning steps at $512 \times 512$ (batch size 4 per GPU). We set $M = 9$ for all reported results. At inference, $10\times$ densification of a full stack takes 0.8 seconds on a single A6000. Additional hyperparameters and architectural details are provided in the Appendix.

%% file: sections/4-results_v2.tex
\subsection{Results}\label{sec:results}

We compare CD-RCM against four conventional interpolation methods: bilinear, B-spline, cubic spline, and Gaussian interpolation, and two learned baseline families: a NeRF-like coordinate MLP and FILM~\citep{reda2022film}. We evaluate photometric fidelity using PSNR and SSIM and perceptual similarity using LPIPS and $\mathcal{L}_{SPF}$. Baseline and evaluation details are provided in the Appendix.

\begin{table}[h]
\footnotesize
    \centering
    \begin{tabular}{l|c|c|c|c}
    \toprule
        \textbf{Method} & PSNR $\uparrow$ & SSIM $\uparrow$ & LPIPS $\downarrow$ & $\mathcal{L}_\text{SPF}$ $\downarrow$ \\
        \midrule
        Bilinear Interpolation &  21.98 & 0.471 & 0.377 & \underline{0.153} \\
        B-Spline Interpolation  & 22.02 & 0.471 & \underline{0.375} & 0.161 \\
        Cubic Spline Interpolation  & 22.09 & 0.469 & 0.378 & 0.163 \\
        Gaussian Interpolation  & \underline{22.87} & \underline{0.522} & 0.477 & 0.164 \\
        \midrule
        MLP (NeRF) & 19.10 & 0.445 & 0.520 & 0.480 \\
        FiLM zero-shot (pretrained) & 20.24 & 0.380 & 0.440 & 0.156 \\
        FiLM fine-tuned (5 epochs) & 22.83 & 0.510 & 0.496 & 0.181 \\
        \midrule
        \textbf{CD-RCM} (Ours)  &  \textbf{23.36} & \textbf{0.581} & \textbf{0.314} & \textbf{0.145} \\
        \bottomrule
    \end{tabular}
    \vspace{2mm}
    \caption{\textbf{Quantitative comparison.} 
    CD-RCM outperforms classical interpolation and learned baselines across photometric (PSNR, SSIM) and perceptual (LPIPS, $\mathcal{L}_{SPF}$) metrics.
    }
    \vspace{-3mm}
    \label{tab:quant-results}
\end{table}

\begin{figure}[h]
    \centering
    \vspace{-2mm}
    \includegraphics[width=\linewidth]{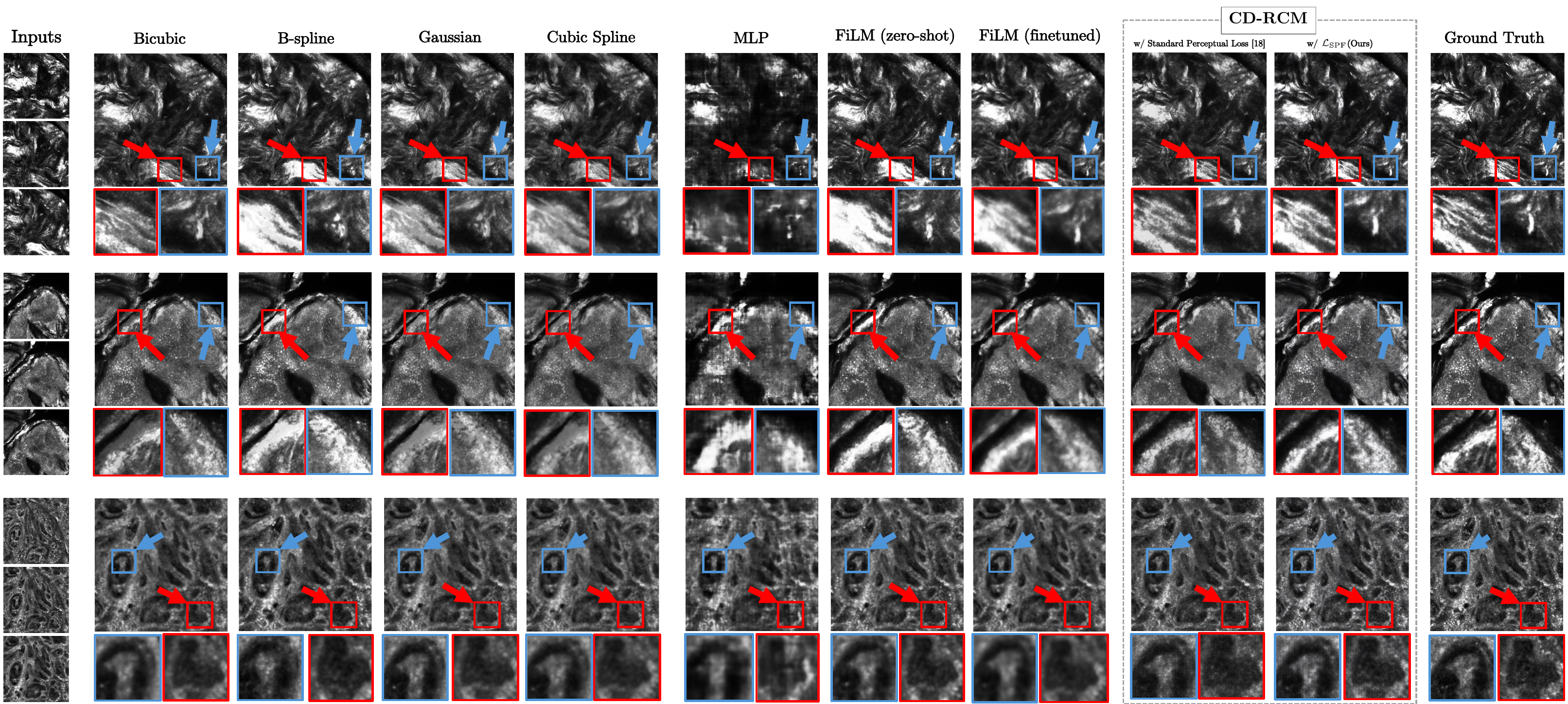}
    \caption{\textbf{Qualitative comparison of novel-depth synthesis methods.} We compare CD-RCM with classical interpolation and learned baselines. CD-RCM better preserves fine cellular structures and depth-dependent morphology, producing predictions closest to the ground truth. Insets highlight representative structural differences and reconstruction artifacts.}
    \label{fig:qual-results}
\end{figure}

As shown in \cref{tab:quant-results,fig:qual-results}, CD-RCM achieves the best performance across all quantitative metrics while producing reconstructions that most closely preserve the ground-truth tissue structure. Clinicians generally prefer reconstructions that preserve high-frequency details, even with mild noise, over overly smoothed images that obscure diagnostically relevant structures. Among classical baselines, Gaussian interpolation introduces noticeable blur. Although this smoothing can improve PSNR, which favors reduced variance, it degrades perceptual fidelity and obscures diagnostically relevant detail. The per-stack NeRF-like coordinate MLP performs poorly despite accurately fitting the observed slices, indicating that photometric optimization without a cross-stack anatomical prior is insufficient for reconstructing unseen tissue structure.

FiLM~\citep{reda2022film} exhibits a similar tradeoff between photometric and perceptual fidelity. Zero-shot FILM produces relatively sharp predictions but introduces inter-slice discontinuities and morphologically implausible structures. Fine-tuning FiLM on RCM substantially improves PSNR and SSIM by producing smoother predictions, but suppresses fine morphological detail and worsens both perceptual metrics. In contrast, CD-RCM preserves layered tissue structure and fine cellular features while maintaining continuity across depth, achieving the strongest photometric and perceptual fidelity.

\begin{figure}[h]
    \centering
    \includegraphics[width=\linewidth]{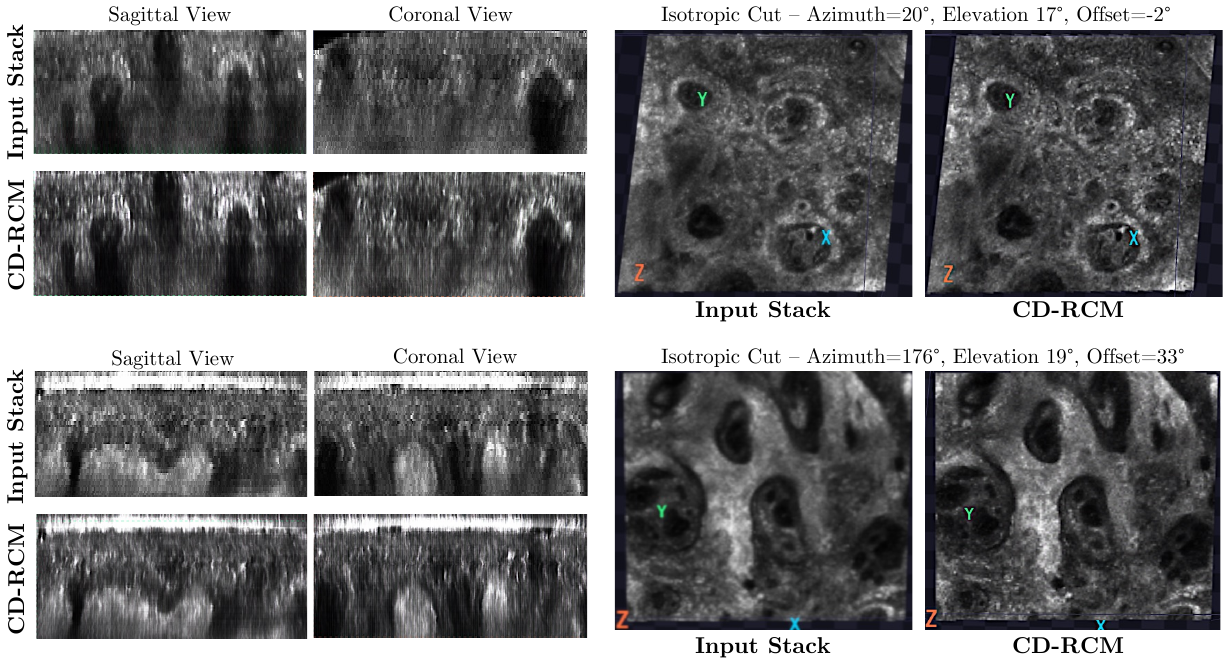}
    \caption{\textbf{Cross-sectional and arbitrary-plane visualization of densified stacks.} 
    We densify the input stacks using CD-RCM to match the axial and lateral resolution of the actual sample, enabling isotropic 3D visualization. Sagittal, coronal, and oblique cross-sections show markedly smoother structural transitions compared to the sparsely sampled input stacks.}
    \vspace{-4mm}
    \label{fig:isotropic-cuts}
\end{figure}

\paragraph{Cross-sectional views of densified stacks}
\cref{fig:isotropic-cuts} shows the cross-sectional and arbitrary-plane visualization of CD-RCM--densified stacks. Using CD-RCM, we densify RCM stacks so that axial sampling matches the actual sample dimensions, enabling isotropic 3D visualization. We show sagittal and coronal cross-sections (left) and arbitrary oblique cuts at specified azimuth, elevation, and offset angles (right) for two example stacks. Compared to the sparsely sampled input stacks, CD-RCM produces markedly smoother structural transitions along the
$z$-axis, reducing the staircase artifacts visible in the original cross-sections and revealing continuous tissue morphology that may facilitate identification of clinically relevant landmarks.

%% file: sections/5-ablations_v2.tex
\subsection{Ablation Study}\label{sec:ablation}
We ablate key design choices of CD-RCM, mainly the training objective and the design of our skin-specific perceptual loss $\mathcal{L}_\text{SPF}$. All models are trained at $256\times256$ resolution for 20k steps.

\begin{table}[]
\footnotesize
    \centering
    \begin{tabular}{l|c|c|c|c}
    \toprule
        \textbf{Training objective} & PSNR $\uparrow$ & SSIM $\uparrow$ & LPIPS $\downarrow$ & $\mathcal{L}_\text{SPF}$ $\downarrow$ \\
        \midrule
        ($\mathcal{L}_\text{MSE}$)  &  &  & & \\
        \hspace{0.5mm} + $\mathcal{L}_\text{LPIPS}$   & 21.60 & 0.480 & 0.343 & 0.273 \\
        \midrule
        \hspace{0.5mm} + Perceptual Loss~\citep{johnson2016perceptual}  & 22.56 & 0.547 & 0.386 & 0.258 \\
        \hspace{0.5mm} + $\mathcal{L}_\text{LPIPS}$ + Perceptual Loss~\citep{johnson2016perceptual}  & \underline{22.65} & \underline{0.567} & \underline{0.329} & \underline{0.216} \\
        \midrule
        \hspace{0.5mm} + $\mathcal{L}_\text{LPIPS}$ + $\mathcal{L}_\text{SPF}$ (multi-feature)  & 22.39 & 0.522 & 0.338 & 0.246 \\
        \hspace{0.5mm} + $\mathcal{L}_\text{LPIPS}$ + $\mathcal{L}_\text{SPF}$(\cref{eq:loss-spf}) & \textbf{23.35} & \textbf{0.597} & \textbf{0.288} & \textbf{0.172} \\
        \bottomrule
    \end{tabular}
    \vspace{2mm}
    \caption{\textbf{Ablation of training objectives.} We compare different perceptual loss combinations for training CD-RCM. Best and second-best results are \textbf{bolded} and \underline{underlined}, respectively. Our proposed $\mathcal{L}_\text{SPF}$ (\cref{eq:loss-spf}) yields the best performance across all metrics.}
    \vspace{-5mm}
    \label{tab:ablation}
\end{table}

\paragraph{Training Objective.}
\cref{tab:ablation} reports performance across perceptual loss configurations, with $\mathcal{L}_\text{MSE}$ included in all experiments. LPIPS alone yields the weakest results. The VGG-based perceptual loss of~\cite{johnson2016perceptual} performs better, and combining it with LPIPS further boosts performance. However, replacing the VGG-based loss with our proposed $\mathcal{L}_\text{SPF}$ (\cref{eq:loss-spf}) yields the best overall performance, simultaneously improving all four metrics. \cref{fig:qual-results} (columns 5 and 6) qualitatively confirms that $\mathcal{L}_\text{SPF}$ produces higher-fidelity reconstructions that more closely match the ground truth.

\paragraph{Design of $\mathcal{L}_\text{SPF}$.}
We also ablate the design of $\mathcal{L}_\text{SPF}$ by comparing our final-layer MAE formulation (\cref{eq:loss-spf}) against a multi-layer feature variant analogous to the standard perceptual loss~\citep{johnson2016perceptual}. As shown in the last two rows of \cref{tab:ablation}, the single-layer design outperforms the multi-layer alternative across all metrics.

CD-RCM demonstrates that feedforward novel-depth synthesis can efficiently densify sparse RCM stacks without per-stack optimization, producing anatomically plausible intermediate slices in a single forward pass. More broadly, our results highlight the potential of adapting neural rendering methods to domain-specific imaging physics in computational microscopy. A promising direction is to accelerate stack acquisition by sparsely sampling tissue sections and reconstructing intermediate slices with CD-RCM. This could speed up clinical imaging and reduce motion artifacts by shortening the time patients must remain still (see Limitations on registration).

%% file: sections/6-expert-eval.tex
\subsection{Expert Evaluation}
\label{sec:expert_eval}

Image-similarity metrics do not fully capture morphological fidelity, particularly when smoothing improves PSNR while suppressing fine tissue structure. We therefore conducted a blinded evaluation with an RCM expert. Images were anonymized and presented in randomized order. The study comprised two complementary tasks, namely:

\begin{itemize}[leftmargin=1em]

\item{\textbf{Image quality.}
The reader independently evaluated 25 en face slices: five acquired slices and five each from CD-RCM, fine-tuned FILM, Gaussian, and B-spline interpolation. For each slice, the reader rated 1--3 its suitability for evaluating tissue morphology, presence of morphologically implausible structures or artifacts, and confidence in interpretation.}

\item{\textbf{Similarity to ground truth.}
For 25 additional cases, the reader was shown two acquired slices surrounding a withheld ground-truth slice and reconstructions from CD-RCM, fine-tuned FILM, Gaussian, and B-Spline interpolation in randomized order. The reader selected the reconstruction most similar to the ground truth. To evaluate generalization beyond the training spacing, inputs were separated by 6 slices ($9\,\mu$m), $1.5\times$ the spacing used during training, with the target midway between them.}

\end{itemize}

\begin{table}[t]
  \centering
  \footnotesize
  \begin{tabular}{lccc}
    \toprule
    \textbf{Method} & \textbf{Suitability}$\uparrow$ & \textbf{Implausibility}$\downarrow$ & \textbf{Confidence}$\uparrow$ \\
    \midrule
    Acquired (ground truth) & 2.40 (0.89) & 1.80 (0.84) & 2.40 (0.89) \\
    \midrule
    B-Spline Interpolation  & 2.60 (0.55) & 1.80 (0.84) & 2.80 (0.45) \\
    Gaussian Interpolation  & 1.60 (0.55) & 2.00 (0.71) & 2.00 (0.71) \\
    FILM fine-tuned         & 1.80 (0.45) & 2.00 (0.71) & 2.00 (0.71) \\
    CD-RCM (Ours)           & 2.40 (0.55) & 1.60 (0.55) & 2.60 (0.55) \\
    \bottomrule
  \end{tabular}
  \vspace{3pt}
  \caption{\textbf{Expert evaluation of image quality.} For 5 images per method, we report the blinded ratings of individual slices on a 1-3 scale. CD-RCM was rated comparably to ground-truth slices.}
  \label{tab:expert_eval-1}
\end{table}

\begin{table}[t]
  \centering
  \footnotesize
  \begin{tabular}{lcc} 
    \toprule
    Method & Times chosen & Rate \\
    \midrule
    B-Spline Interpolation   & 3  & 12.0\% \\
    Gaussian Interpolation & 0  & 0.0\%  \\
    FILM fine-tuned        & 0  & 0.0\%  \\
    CD-RCM (Ours)          & 22 & \textbf{88.0\%} \\
    \bottomrule
  \end{tabular}
  \vspace{3pt}
  \caption{\textbf{Expert evaluation of similarity to ground truth.} For  25 cases, we report the frequency with which each reconstruction was judged most similar to the withheld ground truth. CD-RCM was rated most similar in the majority of cases.}
  \vspace{-5mm}
  \label{tab:expert_eval-2}
\end{table}

As shown in \cref{tab:expert_eval-1}, CD-RCM received ratings comparable to acquired slices for suitability (2.40 vs.\ 2.40), implausibility (1.60 vs.\ 1.80), and confidence (2.60 vs.\ 2.40). Gaussian interpolation and fine-tuned FILM received lower suitability and confidence ratings, consistent with the loss of fine morphological detail from smoothing observed in \cref{sec:results}. Given the single reader and small sample size, we treat these ratings as indicative rather than as a formal non-inferiority test.

In the ground-truth comparison reported in \cref{tab:expert_eval-2}, CD-RCM was selected in 22/25 cases, compared with 3/25 for linear interpolation and none for FILM or Gaussian interpolation. 
Notably, FILM achieved the highest PSNR in 9/25 cases, whereas CD-RCM achieved the lowest LPIPS in all 25, further illustrating that expert judgments aligned better with perceptual than pixel-wise similarity.

%% file: sections/7-limitations-and-discussion.tex
\vspace{-2mm}
\subsection{Limitations}\label{sec:limitations}
\vspace{-2mm}
Our pipeline uses SIFT-based registration to reduce inter-slice motion, but residual misalignment persists in some stacks and can degrade reconstruction quality. Moreover, curating a larger and more diverse dataset, spanning a wider range of skin types, and filtering out slices with low signal content may further improve robustness and performance.

\vspace{-2mm}
\subsection{Future Work}\label{sec:future-work}
\vspace{-2mm}
This work represents a first-of-its-kind, proof-of-concept study for continuous-depth reconstruction from sparse RCM stacks. Our goal is to motivate further development toward clinical applications, including larger-scale data collection and comprehensive human studies.

Future work will integrate CD-RCM into interactive clinical visualization tools to evaluate whether densified stacks improve downstream analyses, such as DEJ localization and strata delineation. We also aim to extend CD-RCM to RCM mosaics acquired at consecutive depths to enable dense visualization over larger tissue regions.
Finally, adapting our framework to other optical sectioning modalities, such as line-field confocal OCT (LC-OCT), could broaden  its impact beyond RCM. 


%% file: sections/8-conclusion.tex
\section{Conclusion}
We introduced CD-RCM, the first feedforward framework for continuous-depth novel-view synthesis tailored to reflectance confocal microscopy. By modeling RCM acquisition as axial translation and using a decoder-only transformer, CD-RCM synthesizes anatomically plausible en-face slices at arbitrary depths without per-stack optimization and with sub-second inference. We further introduced a skin-specific perceptual loss adapted to the micro-textural patterns of RCM. Quantitatively, CD-RCM outperforms interpolation baselines across photometric and perceptual metrics, with ablations supporting the proposed design choices. A blinded expert evaluation further supports the morphological fidelity of the synthesized slices, with CD-RCM rated comparably to acquired RCM images and most similar to withheld ground truth. By enabling axial densification and arbitrary-orientation virtual sectioning, CD-RCM takes a step toward making continuous 3D visualization of \emph{in vivo} skin a practical clinical tool.

%% file: supp-sections/9-additional-implementation-details.tex
\section{Further Implementation Details}\label{sec:supp-implementation}

\subsection{Skin-specific perceptual feature extractor}\label{sec:supp-spf}
For training the skin-specific perceptual feature extractor $\phi_s$ (Sec. 3.5), we use LoRA with rank 16 and optimize using AdamW~\cite{loshchilov2017decoupled} with a constant learning rate of $1\times10^{-4}$. The teacher network is updated using EMA with momentum $m=0.9995$. The projection MLP $h$ has a bottleneck dimension of 256 and an output dimension of 4096. The teacher and student temperatures in Eqs. 9 and 10
are set to $\tau_t=0.04$ and $\tau_s=0.1$, respectively. We train the model for 4 epochs using one NVIDIA A6000 GPU, which takes 3 days.

\subsection{CD-RCM}\label{sec:supp-cdrcm}
For training CD-RCM (Sec. 3.6), we tokenize images and Pl\"ucker embeddings using a patch size $p=8$, and set the embedding dimension (Eqs. 6 and 7) $d=768$ of the linear projectors. For optimization, we use AdamW~\cite{loshchilov2017decoupled} with $\beta_1=0.9$, $\beta_2=0.95$, and a weight decay of $0.05$ applied to all parameters except LayerNorm layers. We employ a cosine learning rate schedule with linear warmup, using 2000 warmup steps and a peak learning rate of $4\times10^{-4}$ for both training resolutions. The loss weights in Eq. 12 are set to $\lambda = 0.5$ and $\gamma = 0.05$. The full model has 170.8M trainable parameters. We train CD-RCM using four NVIDIA A6000 GPUs. The first stage of training (\ie at 256 resolution) takes 2 days, and finetuning at 512 resolution takes another 2 days.

Following~\cite{jin2025lvsm}, we use gradient skipping for gradient norm $> 5.0$ and gradient-clipping at 1.0 to stabilize training. Lastly, to boost training and inference efficiency, we use FlashAttention-v2~\cite{dao2022flashattention} within the attention blocks, gradient checkpointing~\cite{chen2016training}, and mixed-precision training with BF16 data type.

%% file: supp-sections/10-dataset-audit.tex
\section{Dataset Overlap Audit}
\label{sec:dataset_overlap}

As stated in \cref{sec:experiments}, subject identifiers were not retained with the RCM stacks. Therefore, although the train/test split was performed at the stack level, we cannot guarantee that different skin sites from the same volunteer do not appear across both splits. We therefore conducted a post hoc similarity analysis to assess whether duplicate, adjacent, or overlapping acquisitions of the same skin site were present across the training and evaluation sets.

We embedded all training and evaluation slices using the skin-specific feature extractor from \cref{sec:loss} and computed nearest-neighbor cosine similarity for each of the 60 evaluation stacks against the training set. One evaluation stack had very high similarity ($0.98$) to a training stack, and visual inspection showed matching ring and cellular structures under a lateral shift, consistent with an overlapping acquisition site. Seven additional stacks had similarities of $0.75$-$0.85$, but no corresponding anatomical structures were observed, suggesting feature-space similarity rather than true overlap.

To assess the effect of these potentially overlapping cases, we repeated the aggregate evaluation after excluding these 8 identified stacks. Performance on the remaining 52 evaluation stacks was essentially unchanged, and the excluded stacks did not show anomalously high reconstruction performance relative to the full evaluation set. Thus, while the absence of subject identifiers prevents us from establishing a strictly subject-disjoint split, this analysis indicates that detectable same-site or near-duplicate overlap does not account for the reported performance.

%% file: supp-sections/11-additional-ablation.tex
```latex
\section{Additional Ablation}
\paragraph{Positional encoding.}
CD-RCM conditions each token on the Plücker ray map $\mathbf{P}_i$ (\cref{eq:token-ip,eq:token-tar}). Because the virtual cameras differ only by axial translation, $\mathbf{P}_i$ is fully determined by pixel position $(u,v)$ and canonicalized depth $z_i$. To test whether this representation is important, we replace $\mathbf{P}_i$ with a learned multiresolution hash encoding~\citep{muller2022instant} of $(u,v,z_i)$ while keeping the remaining architecture and training setup fixed.

We normalize $(u,v,z_i)$ to $[0,1]^3$ and use a 16-level hash grid with resolutions increasing geometrically from 16 to 256 and two features per level, yielding a 32-channel positional map. The hash-grid parameters are learned jointly with the transformer.

Both variants are trained on $256\times256$ inputs using the full objective
($\mathcal{L}_{\text{MSE}}+\mathcal{L}_{\text{LPIPS}}+\mathcal{L}_{\text{SPF}}$) and evaluated after 10k steps of the same 20k-step schedule. As shown in Table~\ref{tab:ablation_pe}, the two encodings perform comparably overall: the hash encoding slightly improves PSNR and SSIM, while Plücker rays achieve better perceptual scores. This suggests that CD-RCM does not depend critically on Plücker coordinates, and that alternative continuous depth encodings can provide similar performance.

\begin{table}[t]
  \centering
  \small
  \begin{tabular}{lcccc}
    \toprule
    Positional encoding & PSNR$\uparrow$ & SSIM$\uparrow$ & LPIPS$\downarrow$ & $\mathcal{L}_{\text{SPF}}\downarrow$ \\
    \midrule
    Plücker ray map (default)                      & 22.82          & 0.574          & \textbf{0.281} & \textbf{0.156} \\
    Multiresolution hash~\citep{muller2022instant} & \textbf{23.10} & \textbf{0.605} & 0.298          & 0.187 \\
    \bottomrule
  \end{tabular}
  \vspace{3mm}
  \caption{\textbf{Ablation of positional encoding.} Both models are evaluated after 10k training steps on 300 windows from the 60 evaluation stacks; only the encoding of $(u,v,z)$ differs.}
  \label{tab:ablation_pe}
\end{table}

%% file: supp-sections/12-eval-baselines.tex
\section{Baselines Methods}
We adopt four well-established classical interpolation techniques widely adopted for processing medical imaging data as the baselines in our experiments~\cite{enjilela2019cubic,lehmann2002survey}. For completeness, we also evaluate two learning-based baselines: a per-stack coordinate MLP, representative of NeRF-style per-scene optimization~\citep{mildenhall2021nerf}, and FILM~\citep{reda2022film}, a state-of-the-art video frame interpolation network, which we evaluate both zero-shot and after fine-tuning on our training stacks.

We additionally evaluated a diffusion-based model for intermediate slice prediction, but it failed to reconstruct high-frequency structures. This is consistent with the known data requirements of diffusion models and their sensitivity to noisy inputs, which are characteristic of RCM imagery.

Below, we explain each baseline:

\subsection{Bilinear Interpolation}

Bilinear interpolation estimates each intermediate slice as a per-pixel linear blend of the two acquired slices that bracket the target depth. Since the slices of a registered stack are laterally aligned, no in-plane resampling is required, and the method reduces to linear interpolation along the axial direction. Given the bracketing slices $I(z_a, x, y)$ and $I(z_b, x, y)$ with $z_a < z < z_b$, the interpolated intensity at depth $z$ is
\begin{equation}
  \hat{I}(z, x, y) = (1 - \tau)\, I(z_a, x, y) + \tau\, I(z_b, x, y),
  \qquad \tau = \frac{z - z_a}{z_b - z_a}.
\end{equation}

\subsection{B-Spline Interpolation.}
B-spline interpolation models the intensity along the axial direction as a smooth spline function and evaluates it at intermediate depths. For our case, it treats the three input slices as samples of a smooth intensity function along the axial direction and estimates missing slices by applying third-order spline interpolation independently at each pixel location.
For each pixel location $(x,y)$, let the observed slices be $I(z_i,x,y)$ for $i \in \{1,\dots,N\}$. 
The interpolated intensity at depth $z$ is computed as

\begin{equation}
\hat{I}(z,x,y) = \sum_{k} c_k(x,y)\, \beta_3(z-k),
\end{equation}

where $\beta_3(\cdot)$ denotes the cubic B-spline basis function and $c_k(x,y)$ are spline coefficients derived from the observed samples. 
In practice, we implement this interpolation using \texttt{map\_coordinates} of \texttt{scipy.ndimage} library with spline order $3$, which evaluates the cubic B-spline along the depth axis independently for each pixel.

\subsection{Cubic Spline Interpolation.}
Cubic spline interpolation fits a natural cubic spline along the depth axis for each pixel independently, using the three observed slices as control points. 
Given observed samples $(z_i, I(z_i,x,y))$, a piecewise cubic polynomial is constructed, such that

\begin{equation}
S(z) =
\begin{cases}
a_1 z^3 + b_1 z^2 + c_1 z + d_1, & z_1 \le z < z_2 \\
\vdots \\
a_{N-1} z^3 + b_{N-1} z^2 + c_{N-1} z + d_{N-1}, & z_{N-1} \le z \le z_N
\end{cases}
\end{equation}

such that

\begin{equation}
S(z_i) = I(z_i,x,y), \quad
S'(z_i^-)=S'(z_i^+), \quad
S''(z_i^-)=S''(z_i^+),
\end{equation}

with natural boundary conditions

\begin{equation}
S''(z_1)=S''(z_N)=0.
\label{eq:second-der}
\end{equation}

Intermediate slices are obtained by evaluating $S(z)$ at the desired depths $z$.

We implement this method using \texttt{CubicSpline} function of \texttt{scipy.interpolate} with boundary condition set to \texttt{natural}, \ie, the second derivative at curve ends are zero (\cref{eq:second-der}).

\subsection{Gaussian Interpolation.}
Gaussian interpolation first generates intermediate slices using cubic B-spline interpolation and then applies spatial Gaussian smoothing to each synthesized slice. 
Let $\tilde{I}(z,x,y)$ denote the B-spline interpolated slice. 
The final result is obtained by convolving each slice with a Gaussian kernel:

\begin{equation}
\hat{I}(z,x,y) =
(\tilde{I} * G_\sigma)(x,y)
=
\sum_{u,v} \tilde{I}(z,x-u,y-v)\, G_\sigma(u,v),
\end{equation}

where

\begin{equation}
G_\sigma(u,v)=\frac{1}{2\pi\sigma^2}
\exp\left(-\frac{u^2+v^2}{2\sigma^2}\right).
\end{equation}

We use $\sigma=1$ in our implementation. This smoothing produces visually smooth transitions across depths but suppresses high-frequency cellular structures.

\subsection{Coordinate MLP (NeRF-like).}
As a per-scene optimization baseline, we fit a coordinate MLP $f_\theta : (x, y, z) \mapsto I$ independently to each evaluation stack, in the spirit of NeRF~\citep{mildenhall2021nerf}. Since our virtual camera undergoes a pure axial translation without parallax (Section~3.2), each observed pixel directly samples the reflectance at $(x, y, z)$ rather than integrating it along a ray through an occluding volume; we therefore omit volume rendering and regress the intensity directly. All coordinates are normalized to $[-1, 1]$, with $z$ normalized over the depth range of the input slices, and are mapped through the positional encoding of NeRF with $L = 10$ frequency bands,
\begin{equation}
  \gamma(p) = \big(p, \sin(2^0 \pi p), \cos(2^0 \pi p), \ldots,
  \sin(2^{L-1} \pi p), \cos(2^{L-1} \pi p)\big),
\end{equation}
applied to each coordinate separately. The MLP has four hidden layers of 256 units with ReLU activations, followed by a sigmoid output. For each stack, we optimize it with an MSE loss on the pixels of the three input slices only, using Adam with a learning rate of $2\times10^{-3}$ for 1{,}500 iterations of 8{,}192 randomly sampled pixels, and then query it at the target depths. We use one window of three inputs and four targets per evaluation stack, sampled as for CD-RCM. The MLP thus has no information beyond the three inputs of the stack it is fit to, which isolates the contribution of the cross-stack prior learned by CD-RCM.

\subsection{FILM.}
FILM~\citep{reda2022film} synthesizes a frame at an arbitrary time $t \in (0, 1)$ between two input frames using a shared multi-scale feature extractor, bidirectional flow estimation, and a fusion decoder. Since FILM operates on pairs of RGB frames, we replicate each grayscale slice to three channels, pad it to a multiple of 64 pixels, and condition the model on the two acquired slices that bracket the target depth, setting $t$ to the relative depth of the target between them. We tile each evaluation stack into non-overlapping windows of five consecutive slices, use the first and last slices as inputs, and predict the three intermediate slices at $t \in \{0.25, 0.5, 0.75\}$, yielding 643 windows over the 60 evaluation stacks.

\textbf{Zero-shot.} We apply the pretrained checkpoint directly, without any training on RCM data.

\textbf{Fine-tuned.} We fine-tune all parameters of FILM on our 156 training stacks, tiled into five-slice windows in the same way (1{,}581 windows, \textit{i.e.,} 4{,}743 training samples per epoch, as each intermediate slice is a separate sample). Instead of FILM's original combination of $L_1$, perceptual, and style losses, we use $L_1 + 0.5\cdot\mathcal{L}_{\text{LPIPS}}$, matching the photometric--perceptual objective of CD-RCM. We optimize with AdamW~\citep{loshchilov2017decoupled} at a learning rate of $2\times10^{-5}$, using 100 linear warmup steps followed by cosine decay, and a batch size of 9 for 5 epochs.

%% file: supp-sections/13-eval-metrics.tex
\section{Evaluation Metrics}

We evaluate reconstruction quality using both photometric and perceptual metrics. Specifically, we report peak signal-to-noise ratio (PSNR), structural similarity (SSIM), and the learned perceptual image patch similarity (LPIPS). We also use our proposed skin-specific perceptual feature loss $\mathcal{L}_\text{SPF}$ (Eq. 13) as a skin domain-aware perceptual similarity metric.

\subsection{PSNR}
Peak signal-to-noise ratio (PSNR) measures pixel-wise reconstruction fidelity between a predicted image $\hat{I}$ and the ground truth image $I$. It is defined as:

\begin{equation}
\text{PSNR} = 10 \log_{10} \left( \frac{\text{MAX
}^2}{\text{MSE}} \right),
\end{equation}

where $\text{MAX}$ denotes the maximum possible pixel intensity (\ie 255 in our case), and MSE is the mean squared error:

\begin{equation}
\text{MSE} = \frac{1}{HW} \sum_{x,y} (I(x,y) - \hat{I}(x,y))^2 .
\end{equation}

Higher PSNR values indicate better reconstruction fidelity.

\subsection{SSIM}
Structural Similarity Index (SSIM) evaluates perceptual image quality by comparing luminance, contrast, and structural information between two images. It is defined as:

\begin{equation}
\text{SSIM}(I,\hat{I}) =
\frac{(2\mu_I \mu_{\hat{I}} + C_1)(2\sigma_{I\hat{I}} + C_2)}
{(\mu_I^2 + \mu_{\hat{I}}^2 + C_1)(\sigma_I^2 + \sigma_{\hat{I}}^2 + C_2)},
\end{equation}

where $\mu$, $\sigma^2$, and $\sigma_{I\hat{I}}$ denote the mean, variance, and covariance of the images, respectively, and $C_1$, $C_2$ are constants used for numerical stability. SSIM ranges from $-1$ to $1$, with higher values indicating greater structural similarity.

\subsection{LPIPS}
Learned Perceptual Image Patch Similarity (LPIPS) \cite{zhang2018unreasonable} measures perceptual similarity in the feature space of a pretrained convolutional network (we use the VGG backbone in our implementation, as it is shown in the original paper to be the strongest variant). Given feature activations $\phi_l(\cdot)$ at layer $l$, LPIPS is computed as

\begin{equation}
\text{LPIPS}(I,\hat{I}) =
\sum_l w_l \, \| \phi_l(I) - \phi_l(\hat{I}) \|_2^2,
\end{equation}

where $w_l$ are learned channel-wise weights. Unlike pixel-based metrics, LPIPS better captures perceptual differences in texture and structure. Lower LPIPS values indicate greater perceptual similarity.

%% file: supp-sections/14-additional-qual-results.tex



\section{Broader Impacts}\label{sec:broader-impact}
This work has potential positive societal impact in biomedical imaging and clinical workflows. By enabling continuous-depth reconstruction from sparse RCM stacks, CD-RCM may improve visualization, reduce acquisition burden, and support more consistent analysis of skin microstructure. In the long term, such tools could assist clinicians in diagnosis and monitoring, particularly in settings where dense sampling is time-consuming or impractical.

Potential negative impacts may arise from over-reliance on synthesized data. As CD-RCM generates intermediate slices that are not directly observed, there is a risk that inaccuracies or hallucinated structures could be misinterpreted as real tissue features, potentially affecting clinical decisions if used without appropriate validation. To mitigate these risks, CD-RCM should be used with clear communication that synthesized slices are model predictions.

Additionally, the current model is a proof-of-concept study and is trained on a limited dataset of healthy skin from a single imaging setup. To broaden generalization across populations, devices, or pathological conditions, more exhaustive, diverse data should be collected for training and evaluating the model, incorporating uncertainty estimation, and validating the method in downstream clinical tasks before deployment in real-world settings. The same is discussed in \cref{sec:future-work}.

%% file: main.bib
@String(CVPR  = {IEEE Conf. Comput. Vis. Pattern Recog.})

@String(ECCV  = {Eur. Conf. Comput. Vis.})

@String(NeurIPS = {Adv. Neural Inform. Process. Syst.})

@String(ICLR  = {Int. Conf. Learn. Represent.})

@String(TOG   = {ACM Trans. Graph.})

@String(CVPR  = {CVPR})

@String(ECCV  = {ECCV})

@String(NeurIPS = {NeurIPS})

@String(ICLR  = {ICLR})

@String(TOG   = {ACM TOG})

@book{gonzalez2008reflectance,
  title     = {Reflectance Confocal Microscopy of Cutaneous Tumors: An Atlas with Clinical, Dermoscopic and Histological Correlations},
  editor    = {Gonzalez, Salvador and Gill, M. and Halpern, Allan C.},
  edition   = {1},
  year      = {2008},
  publisher = {CRC Press},
  doi       = {10.3109/9780203091562}
}

@Article{jcm14165779,
AUTHOR = {Wojarska, Monika and Kokot, Klaudia and Bernecka, Paulina and Domańska, Natalia and Libik, Agata and Bunevich, Dana and Nowakowska, Dominika and Dzido, Magdalena and Borzyszkowska, Wiktoria and Kazimierczak, Wojciech and Jankau, Jerzy},
TITLE = {In Vivo Confocal Microscopy in the Surgical Treatment of Keratinocyte Carcinomas: A Systematic Review},
JOURNAL = {Journal of Clinical Medicine},
VOLUME = {14},
YEAR = {2025},
NUMBER = {16},
ARTICLE-NUMBER = {5779},
URL = {https://www.mdpi.com/2077-0383/14/16/5779},
PubMedID = {40869605},
ISSN = {2077-0383},
DOI = {10.3390/jcm14165779}
}

@article{HUZAIRA2001846,
title = {Topographic Variations in Normal Skin, as Viewed by In Vivo Reflectance Confocal Microscopy},
journal = {Journal of Investigative Dermatology},
volume = {116},
number = {6},
pages = {846-852},
year = {2001},
issn = {0022-202X},
doi = {https://doi.org/10.1046/j.0022-202x.2001.01337.x},
url = {https://www.sciencedirect.com/science/article/pii/S0022202X15412552},
author = {Misbah Huzaira and Francisca Rius and Milind Rajadhyaksha and R. Rox Anderson and Salvador González},
}

@article{bozkurt2021skin,
  title={Skin strata delineation in reflectance confocal microscopy images using recurrent convolutional networks with attention},
  author={Bozkurt, Alican and Kose, Kivanc and Coll-Font, Jaume and Alessi-Fox, Christi and Brooks, Dana H and Dy, Jennifer G and Rajadhyaksha, Milind},
  journal={Scientific reports},
  volume={11},
  number={1},
  pages={12576},
  year={2021},
  publisher={Nature Publishing Group UK London}
}

@article{chibane2020neural,
  title={Neural unsigned distance fields for implicit function learning},
  author={Chibane, Julian and Pons-Moll, Gerard and others},
  journal={Advances in Neural Information Processing Systems},
  volume={33},
  pages={21638--21652},
  year={2020}
}

@inproceedings{mescheder2019occupancy,
  title={Occupancy networks: Learning 3d reconstruction in function space},
  author={Mescheder, Lars and Oechsle, Michael and Niemeyer, Michael and Nowozin, Sebastian and Geiger, Andreas},
  booktitle={Proceedings of the IEEE/CVF conference on computer vision and pattern recognition},
  pages={4460--4470},
  year={2019}
}

@article{ghanta2016marked,
  title={A marked Poisson process driven latent shape model for 3D segmentation of reflectance confocal microscopy image stacks of human skin},
  author={Ghanta, Sindhu and Jordan, Michael I and Kose, Kivanc and Brooks, Dana H and Rajadhyaksha, Milind and Dy, Jennifer G},
  journal={IEEE Transactions on Image Processing},
  volume={26},
  number={1},
  pages={172--184},
  year={2016},
  publisher={IEEE}
}

@article{bozkurt2017unsupervised,
  title={Unsupervised delineation of stratum corneum using reflectance confocal microscopy and spectral clustering},
  author={Bozkurt, A and Kose, K and Alessi-Fox, C and Dy, JG and Brooks, DH and Rajadhyaksha, M},
  journal={Skin Research and Technology},
  volume={23},
  number={2},
  pages={176--185},
  year={2017},
  publisher={Wiley Online Library}
}

@article{edwards2017diagnostic,
  title={Diagnostic accuracy of reflectance confocal microscopy using VivaScope for detecting and monitoring skin lesions: a systematic review},
  author={Edwards, SJ and Osei-Assibey, G and Patalay, R and Wakefield, V and Karner, C},
  journal={Clinical and Experimental Dermatology},
  volume={42},
  number={3},
  pages={266--275},
  year={2017},
  publisher={Blackwell Publishing Ltd Oxford, UK}
}

@article{richarz2022challenges,
  title   = {Challenges for New Adopters in Pre-Surgical Margin Assessment by Handheld Reflectance Confocal Microscope of Basal Cell Carcinoma: A Prospective Single-center Study},
  author  = {Richarz, N. A. and Boada, A. and Jaka, A. and Bassas, J. and Ferr{\'a}ndiz, C. and Carrascosa, J. M. and Y{\'e}lamos, O.},
  journal = {Dermatology Practical \& Conceptual},
  year    = {2022},
  volume  = {12},
  number  = {4},
  pages   = {e2022162},
  month   = {Oct},
  doi     = {10.5826/dpc.1204a162},
  pmid    = {36534521},
  pmcid   = {PMC9681170}
}

@inproceedings{jin2025lvsm,
title={LVSM: A Large View Synthesis Model with Minimal 3D Inductive Bias},
author={Haian Jin and Hanwen Jiang and Hao Tan and Kai Zhang and Sai Bi and Tianyuan Zhang and Fujun Luan and Noah Snavely and Zexiang Xu},
booktitle=ICLR,
year={2025},
}

@article{kerbl20233d,
  title={3d gaussian splatting for real-time radiance field rendering.},
  author={Kerbl, Bernhard and Kopanas, Georgios and Leimk{\"u}hler, Thomas and Drettakis, George},
  journal=TOG,
  volume={42},
  number={4},
  pages={139--1},
  year={2023}
}

@inproceedings{imtiaz2025lvt,
  title={LVT: Large-Scale Scene Reconstruction via Local View Transformers},
  author={Imtiaz, Tooba and Chai, Lucy and Heal, Kathryn and Luo, Xuan and Park, Jungyeon and Dy, Jennifer and Flynn, John},
  booktitle={Proceedings of the SIGGRAPH Asia 2025 Conference Papers},
  pages={1--12},
  year={2025}
}

@inproceedings{zhang2024gs,
  title={Gs-lrm: Large reconstruction model for 3d gaussian splatting},
  author={Zhang, Kai and Bi, Sai and Tan, Hao and Xiangli, Yuanbo and Zhao, Nanxuan and Sunkavalli, Kalyan and Xu, Zexiang},
  booktitle=ECCV,
  pages={1--19},
  year={2024},
  organization={Springer}
}

@inproceedings{charatan2024pixelsplat,
  title={pixelsplat: 3d gaussian splats from image pairs for scalable generalizable 3d reconstruction},
  author={Charatan, David and Li, Sizhe Lester and Tagliasacchi, Andrea and Sitzmann, Vincent},
  booktitle=CVPR,
  pages={19457--19467},
  year={2024}
}

@article{plucker1865xvii,
  title={Xvii. on a new geometry of space},
  author={Pl\"ucker, Julius},
  journal={Philosophical Transactions of the Royal Society of London},
  number={155},
  pages={725--791},
  year={1865},
  publisher={The Royal Society London}
}

@inproceedings{dosovitskiy2020vit,
  title={An Image is Worth 16x16 Words: Transformers for Image Recognition at Scale},
  author={Dosovitskiy, Alexey and Beyer, Lucas and Kolesnikov, Alexander and Weissenborn, Dirk and Zhai, Xiaohua and Unterthiner, Thomas and  Dehghani, Mostafa and Minderer, Matthias and Heigold, Georg and Gelly, Sylvain and Uszkoreit, Jakob and Houlsby, Neil},
  booktitle=ICLR,
  year={2021}
}

@inproceedings{kose2020detection,
  title={Detection of the DEJ and Segmentation of Its Morphological Patterns in RCM Images of Melanocytic Skin Lesions},
  author={Kose, Kivanc and Bozkurt, Alican and Fox, Christi A and Gill, Melissa and Brooks, Dana and Dy, Jennifer and Rajadhyaksha, Milind},
  booktitle={Microscopy Histopathology and Analytics},
  pages={MW2A--1},
  year={2020},
  organization={Optica Publishing Group}
}

@inproceedings{cai2024radiative,
  title={Radiative gaussian splatting for efficient x-ray novel view synthesis},
  author={Cai, Yuanhao and Liang, Yixun and Wang, Jiahao and Wang, Angtian and Zhang, Yulun and Yang, Xiaokang and Zhou, Zongwei and Yuille, Alan},
  booktitle={European Conference on Computer Vision},
  pages={283--299},
  year={2024},
  organization={Springer}
}

@article{li20253dgr,
  title={3DGR-CT: Sparse-view CT reconstruction with a 3D Gaussian representation},
  author={Li, Yingtai and Fu, Xueming and Li, Han and Zhao, Shang and Jin, Ruiyang and Zhou, S Kevin},
  journal={Medical Image Analysis},
  volume={103},
  pages={103585},
  year={2025},
  publisher={Elsevier}
}

@inproceedings{r2_gaussian,
  title={R$^2$-Gaussian: Rectifying Radiative Gaussian Splatting for Tomographic Reconstruction},
  author={Ruyi Zha and Tao Jun Lin and Yuanhao Cai and Jiwen Cao and Yanhao Zhang and Hongdong Li},
  booktitle = {Advances in Neural Information Processing Systems (NeurIPS)},
  year={2024}
}

@inproceedings{johnson2016perceptual,
  title={Perceptual losses for real-time style transfer and super-resolution},
  author={Johnson, Justin and Alahi, Alexandre and Fei-Fei, Li},
  booktitle={European conference on computer vision},
  pages={694--711},
  year={2016},
  organization={Springer}
}

@article{simeoni2025dinov3,
  title={Dinov3},
  author={Sim{\'e}oni, Oriane and Vo, Huy V and Seitzer, Maximilian and Baldassarre, Federico and Oquab, Maxime and Jose, Cijo and Khalidov, Vasil and Szafraniec, Marc and Yi, Seungeun and Ramamonjisoa, Micha{\"e}l and others},
  journal={arXiv preprint arXiv:2508.10104},
  year={2025}
}

@article{hu2022lora,
  title={Lora: Low-rank adaptation of large language models.},
  author={Hu, Edward J and Shen, Yelong and Wallis, Phillip and Allen-Zhu, Zeyuan and Li, Yuanzhi and Wang, Shean and Wang, Liang and Chen, Weizhu and others},
  journal={Iclr},
  volume={1},
  number={2},
  pages={3},
  year={2022}
}

@article{radford2019language,
  title={Language models are unsupervised multitask learners},
  author={Radford, Alec and Wu, Jeffrey and Child, Rewon and Luan, David and Amodei, Dario and Sutskever, Ilya and others},
  journal={OpenAI blog},
  volume={1},
  number={8},
  pages={9},
  year={2019}
}

@inproceedings{zhang2018unreasonable,
  title={The unreasonable effectiveness of deep features as a perceptual metric},
  author={Zhang, Richard and Isola, Phillip and Efros, Alexei A and Shechtman, Eli and Wang, Oliver},
  booktitle=CVPR,
  pages={586--595},
  year={2018}
}

@inproceedings{mildenhall2021nerf,
    author = {Mildenhall, Ben and Srinivasan, Pratul P and Tancik, Matthew and Barron, Jonathan T and Ramamoorthi, Ravi and Ng, Ren},
    title = {Nerf: Representing scenes as neural radiance fields for view synthesis},
    booktitle = ECCV,
    year = {2020},
}

@inproceedings{dao2022flashattention,
  title={Flashattention: Fast and memory-efficient exact attention with io-awareness},
  author={Dao, Tri and Fu, Dan and Ermon, Stefano and Rudra, Atri and R{\'e}, Christopher},
  booktitle=NIPS,
  volume={35},
  pages={16344--16359},
  year={2022}
}

@article{atak2023confocal,
  title={Confocal microscopy for diagnosis and management of cutaneous malignancies: clinical impacts and innovation},
  author={Atak, Mehmet Fatih and Farabi, Banu and Navarrete-Dechent, Cristian and Rubinstein, Gennady and Rajadhyaksha, Milind and Jain, Manu},
  journal={Diagnostics},
  volume={13},
  number={5},
  pages={854},
  year={2023},
  publisher={MDPI}
}

@article{lowe2004distinctive,
  title={Distinctive image features from scale-invariant keypoints},
  author={Lowe, David G},
  journal={International journal of computer vision},
  volume={60},
  number={2},
  pages={91--110},
  year={2004},
  publisher={Springer}
}

@inproceedings{henry2020query,
  title={Query-key normalization for transformers},
  author={Henry, Alex and Dachapally, Prudhvi Raj and Pawar, Shubham Shantaram and Chen, Yuxuan},
  booktitle={Findings of the Association for Computational Linguistics: EMNLP 2020},
  pages={4246--4253},
  year={2020}
}

@article{loshchilov2017decoupled,
  title={Decoupled weight decay regularization},
  author={Loshchilov, Ilya and Hutter, Frank},
  journal={arXiv preprint arXiv:1711.05101},
  year={2017}
}

@article{chen2016training,
  title={Training deep nets with sublinear memory cost},
  author={Chen, Tianqi and Xu, Bing and Zhang, Chiyuan and Guestrin, Carlos},
  journal={arXiv preprint arXiv:1604.06174},
  year={2016}
}

@article{schindelin2012fiji,
  title={Fiji: an open-source platform for biological-image analysis},
  author={Schindelin, Johannes and Arganda-Carreras, Ignacio and Frise, Erwin and Kaynig, Verena and Longair, Mark and Pietzsch, Tobias and Preibisch, Stephan and Rueden, Curtis and Saalfeld, Stephan and Schmid, Benjamin and others},
  journal={Nature methods},
  volume={9},
  number={7},
  pages={676--682},
  year={2012},
  publisher={Nature Publishing Group}
}

@article{RCM_LSM,
author = {Rajadhyaksha, Milind and Marghoob, Ashfaq and Rossi, Anthony and Halpern, Allan C. and Nehal, Kishwer S.},
title = {Reflectance confocal microscopy of skin in vivo: From bench to bedside},
journal = {Lasers in Surgery and Medicine},
volume = {49},
number = {1},
pages = {7-19},
doi = {https://doi.org/10.1002/lsm.22600},
url = {https://onlinelibrary.wiley.com/doi/abs/10.1002/lsm.22600},
eprint = {https://onlinelibrary.wiley.com/doi/pdf/10.1002/lsm.22600},
year = {2017}
}

@article{LightSheetM,
   author = "Liu, Jonathan T.C. and Glaser, Adam K. and Poudel, Chetan and Vaughan, Joshua C.",
   title = "Nondestructive 3D Pathology with Light-Sheet Fluorescence Microscopy for Translational Research and Clinical Assays", 
   journal= "Annual Review of Analytical Chemistry",
   year = "2023",
   volume = "16",
   number = "Volume 16, 2023",
   pages = "231-252",
   doi = "https://doi.org/10.1146/annurev-anchem-091222-092734",
   url = "https://www.annualreviews.org/content/journals/10.1146/annurev-anchem-091222-092734",
   publisher = "Annual Reviews",
   issn = "1936-1335",
   type = "Journal Article",
  }

@Article{LCOCT,
AUTHOR = {Latriglia, Flora and Ogien, Jonas and Tavernier, Clara and Fischman, Sébastien and Suppa, Mariano and Perrot, Jean-Luc and Dubois, Arnaud},
TITLE = {Line-Field Confocal Optical Coherence Tomography (LC-OCT) for Skin Imaging in Dermatology},
JOURNAL = {Life},
VOLUME = {13},
YEAR = {2023},
NUMBER = {12},
ARTICLE-NUMBER = {2268},
URL = {https://www.mdpi.com/2075-1729/13/12/2268},
PubMedID = {38137869},
ISSN = {2075-1729},
DOI = {10.3390/life13122268}
}

@article{3DPath,
title = {Analysis of 3D pathology samples using weakly supervised AI},
journal = {Cell},
volume = {187},
number = {10},
pages = {2502-2520.e17},
year = {2024},
issn = {0092-8674},
doi = {https://doi.org/10.1016/j.cell.2024.03.035},
url = {https://www.sciencedirect.com/science/article/pii/S0092867424003519},
author = {Andrew H. Song and Mane Williams and Drew F.K. Williamson and Sarah S.L. Chow and Guillaume Jaume and Gan Gao and Andrew Zhang and Bowen Chen and Alexander S. Baras and Robert Serafin and Richard Colling and Michelle R. Downes and Xavier Farré and Peter Humphrey and Clare Verrill and Lawrence D. True and Anil V. Parwani and Jonathan T.C. Liu and Faisal Mahmood},
}

@article{enjilela2019cubic,
  title={Cubic-spline interpolation for sparse-view ct image reconstruction with filtered backprojection in dynamic myocardial perfusion imaging},
  author={Enjilela, Esmaeil and Lee, Ting-Yim and Wisenberg, Gerald and Teefy, Patrick and Bagur, Rodrigo and Islam, Ali and Hsieh, Jiang and So, Aaron},
  journal={Tomography},
  volume={5},
  number={3},
  pages={300},
  year={2019}
}

@article{lehmann2002survey,
  title={Survey: Interpolation methods in medical image processing},
  author={Lehmann, Thomas Martin and Gonner, Claudia and Spitzer, Klaus},
  journal={IEEE transactions on medical imaging},
  volume={18},
  number={11},
  pages={1049--1075},
  year={2002},
  publisher={IEEE}
}

@inproceedings{reda2022film,
  title={Film: Frame interpolation for large motion},
  author={Reda, Fitsum and Kontkanen, Janne and Tabellion, Eric and Sun, Deqing and Pantofaru, Caroline and Curless, Brian},
  booktitle={European Conference on Computer Vision},
  pages={250--266},
  year={2022},
  organization={Springer}
}

@article{gambini2024video,
  title={Video frame interpolation neural network for 3D tomography across different length scales},
  author={Gambini, Laura and Gabbett, Cian and Doolan, Luke and Jones, Lewys and Coleman, Jonathan N and Gilligan, Paddy and Sanvito, Stefano},
  journal={Nature Communications},
  volume={15},
  number={1},
  pages={7962},
  year={2024},
  publisher={Nature Publishing Group UK London}
}

@article{cao2024neural,
  title={Neural space--time model for dynamic multi-shot imaging},
  author={Cao, Ruiming and Divekar, Nikita S and Nunez, James K and Upadhyayula, Srigokul and Waller, Laura},
  journal={Nature Methods},
  volume={21},
  number={12},
  pages={2336--2341},
  year={2024},
  publisher={Nature Publishing Group US, New York}
}

@article{kang2026adaptive,
  title={Adaptive optical correction for in vivo two-photon fluorescence microscopy with neural fields},
  author={Kang, Iksung and Kim, Hyeonggeon and Natan, Ryan and Zhang, Qinrong and Yu, Stella X and Ji, Na},
  journal={Nature Methods},
  pages={1--10},
  year={2026},
  publisher={Nature Publishing Group US, New York}
}

@article{muller2022instant,
  title={Instant neural graphics primitives with a multiresolution hash encoding},
  author={M{\"u}ller, Thomas and Evans, Alex and Schied, Christoph and Keller, Alexander},
  journal={ACM transactions on graphics (TOG)},
  volume={41},
  number={4},
  pages={1--15},
  year={2022},
  publisher={ACM, New York, NY, USA}
}
